%% file: paper.tex
\documentclass[runningheads]{llncs}
\usepackage[T1]{fontenc}
\usepackage{graphicx}
\usepackage{todonotes}
\usepackage{amsmath}
\usepackage{siunitx}
\usepackage{cleveref}
\usepackage[inline]{enumitem}
\DeclareSIUnit\ft{ft}
\DeclareSIUnit\px{px}

\usepackage{pgfplots}
\pgfplotsset{compat=1.16}

\usepackage{booktabs}
\newlength{\figheight}
\usepackage{tabularx}
\usepackage{multirow}

\usepackage{url}
\usepackage{xspace}

\makeatletter
\DeclareRobustCommand\onedot{\futurelet\@let@token\@onedot}
\def\@onedot{\ifx\@let@token.\else.\null\fi\xspace}

\def\eg{\emph{e.g}\onedot} 
\def\ie{\emph{i.e}\onedot} 
\def\cf{\emph{c.f}\onedot} 
 
\def\wrt{w.r.t\onedot}

\def\Loss{\mathcal{L}}

\makeatother

\begin{document}
%
%\title{Our Paper for 3D UAV Detection\thanks{Supported by organization x.}}
\title{Long Range Object-Level Monocular Depth Estimation for UAVs}

% AUTHORS AND INSTITUTIONS COMMENTED OUT FOR ANONYMITY
%\author{First Author\inst{1}\orcidID{0000-1111-2222-3333} \and
%Second Author\inst{2,3}\orcidID{1111-2222-3333-4444} \and
%Third Author\inst{3}\orcidID{2222--3333-4444-5555}}
%% First names are abbreviated in the running head.
%% If there are more than two authors, 'et al.' is used.
%%
%\institute{Princeton University, Princeton NJ 08544, USA \and
%Springer Heidelberg, Tiergartenstr. 17, 69121 Heidelberg, Germany

\author{David Silva\inst{1}\orcidID{0000-0002-9002-8693} \and
    Nicolas Jourdan\inst{2}\orcidID{0000-0003-0159-2442} \and
    Nils G\"ahlert\inst{1}\orcidID{0000-0002-3622-2903}}
\authorrunning{D. Silva et al.}
\institute{Wingcopter GmbH, Germany
    \email{\{silva,gaehlert\}@wingcopter.com} \and TU Darmstadt, Germany \email{n.jourdan@ptw.tu-darmstadt.de} }

\titlerunning{Long Range Object-Level MDE for UAVs}
\maketitle              % typeset the header of the contribution
\begin{abstract}
    Computer vision-based object detection is a key modality for advanced Detect-And-Avoid systems that allow for autonomous flight missions of UAVs.
    While standard object detection frameworks do not predict the actual depth of an object, this information is crucial to avoid collisions.
    In this paper, we propose several novel extensions to state-of-the-art methods for monocular object detection from images at long range.
    Firstly, we propose Sigmoid and ReLU-like encodings when modeling depth estimation as a regression task.
    Secondly, we frame the depth estimation as a classification problem and introduce a Soft-Argmax function in the calculation of the training loss.
    The extensions are exemplarily applied to the YOLOX object detection framework.
    We evaluate the performance using the Amazon Airborne Object Tracking dataset.
    In addition, we introduce the Fitness score as a new metric that jointly assesses both object detection and depth estimation performance.
    Our results show that the proposed methods outperform state-of-the-art approaches \wrt existing, as well as the proposed metrics.

\end{abstract}

%
% ---- Body ----
%
\input{chapters/1_introduction}
\input{chapters/2_related_work}
\input{chapters/3_methodology}
\input{chapters/4_experiments}

\input{chapters/5_conclusion}

\clearpage

\bibliographystyle{splncs04}
\bibliography{bibliography}

\end{document}

%% file: chapters/1_introduction.tex
\section{Introduction}
Within recent years, significant technological progress in Unmanned Aerial Vehicles (UAVs) was achieved.
To enable autonomous flight missions and mitigate the risk of in-flight collisions, advanced Detect-And-Avoid (DAA) systems need to be deployed to the aircraft.
By design, these systems shall maintain a \emph{well-clear volume} around other airborne traffic \cite{astmf3442}.
As a result, DAA systems are required to reliably detect potential intruders and other dangerous objects at a long range to allow for sufficient time to plan and execute avoidance maneuvers.
%This requirement can only be achieved by employing a perception stack to reliably detect the objects of interest.
Specifically in small and lightweight UAVs, the usage of Lidar and Radar systems is challenging due to their power consumption, weight, and the required long range detection capabilities.
However, computer vision approaches based on monocular images have proved their effectiveness in related use cases such as autonomous driving \cite{cordts2016cityscapes,janai2020computer,caesar2020nuscenes}.
In addition, cameras can be equipped with lenses that employ different focal lengths depending on the application.
Camera systems are therefore a powerful base modality for the perception stack of small and lightweight UAVs.

Depending on the actual use case and application, engineers might choose from several computer vision-related tasks such as image classification, object detection, or semantic segmentation.
Those tasks are nowadays usually solved by Convolutional Neural Networks (CNNs) specifically designed for the selected use case.
For vision-based DAA systems, single-stage object detection frameworks like SSD \cite{liu2016ssd} or YOLO \cite{redmon2016you,redmon2017yolo9000,yolox2021} are often employed to detect the objects of interest.
By default, these frameworks detect objects in the two-dimensional image space by means of axis-aligned, rectangular bounding boxes.
To reliably detect and avoid potentially hazardous objects, additional information about their three-dimensional position and trajectory is crucial.
This capability, however, is missing in most vanilla object detection frameworks and specific extensions are needed to provide it.

%There are two major challengees in retrieving the three-dimensional position of objects in monocular images:
%Firstly, the increased complexity in detecting objects in 3D space compared to 2D space.
%Secondly, the indistinguishability between distance and scale that naturally appears in monocular images.
% In case of using monocular images as the base modality estimating the accurate depth also renders a challenge that

% While in autonomous driving, most of the objects of interest are quite close, \ie, less then \SI{200}{\meter}, UAVs are required to keep a well-clear distance of at least \SI{2000}{\ft} or approximately \SI{600}{\meter}.

%In addition, monocular depth estimation can be split into two streams: pixel-based depth estmation -- \ie, dense depth estimation -- and depth estimation on an object level.
%Pixel-based depth estimation from monocular images is a well-studied problem in both academia as well as in industrial applications \todo{cite}.
%In contrast, object-based depth estimation from monocular images only gained attention recently.
%Most of the corresponding studies focus on the application of autonomous driving \todo{cite}.
%For UAVs, however, object-basedd depth estimation renders an additional challenge due to the required long range for detection of up to several kilometers that exceeds the typical range used in autonomous driving.

In this paper, we specifically address the problem of object-level depth estimation based on monocular images for long range detections in the use case of UAVs.
Several studies focusing on monocular 3D object detection have been conducted in autonomous driving \cite{janai2020computer,caesar2020nuscenes,sun2020scalability,gahlert2020cityscapes,gahlert2020single}.
For UAV-related use cases, however, object-level monocular depth estimation at long range is not yet widely researched.
The two fields of application, UAVs and autonomous driving, differ in two major aspects:
\begin{enumerate*}
    \item The range of the objects. UAVs are required to keep a well clear volume of at least \SI{2000}{\ft} or approximately \SI{600}{\meter} to prevent potential mid-air collisions \cite{astmf3442}.
    In autonomous driving, on the other hand, objects are mostly limited to less than \SI{200}{\meter} \cite{geiger2012we,gahlert2020cityscapes,sun2020scalability,caesar2020nuscenes,janai2020computer,ye2022rope3d}.
    \item Knowledge of the full 9 degrees of freedom 3D bounding box is not required to maintain the well clear volume. The distance is sufficient. In addition to simplifying the task, this aspect greatly eases the annotation process. As objects do not require a fully annotated 3D bounding box, one can can save both time and money.
\end{enumerate*}

Thus, we summarize our contributions as follows:
\begin{enumerate*}
    \item We propose two encodings, Sigmoid and ReLU-like, to improve long range depth estimation modeled as a regression task.
    \item We frame the task of depth estimation as a classification problem and introduce Soft-Argmax based loss functions to improve the performance of monocular depth estimation.
    \item We introduce a novel \emph{Fitness Score} metric to assess the quality of depth estimation on object-level combined with the object detection metrics.
    \item We demonstrate the extension of the state-of-the-art YOLOX object detection framework and benchmark the proposed approaches against existing methods applied to long range detections.
\end{enumerate*}

%% file: chapters/2_related_work.tex
\section{Related Work}
The problem of depth estimation from monocular RGB images has been the subject of intense academic research in the past decade.
Due to the ambiguity between an object's size and the object's distance to the camera, it is mathematically an ill-posed problem \cite{lee2019big,gahlert2020cityscapes}.
Thus, machine learning approaches, specifically ones that rely on CNNs, gained traction in this field.
Two research streams can be identified in monocular depth estimation:
\begin{enumerate*}
    \item \textit{Dense} or \emph{Pixel-level} depth estimation, which estimates a dense map of distances for every pixel of a given image, and
    \item \textit{Object-level} depth estimation, which estimates distances only for detected objects of interest.
\end{enumerate*}
While dense depth estimation is more prominent in computer vision literature, 3D object detection is gaining popularity in relevant application domains such as environment perception for autonomous driving \cite{janai2020computer,caesar2020nuscenes,sun2020scalability,gahlert2020cityscapes,gahlert2020single}.
Nevertheless, there's limited related work in the domain of 2D object-level depth estimation at long ranges \cite{ghosh2022airtrack}.
%
% The employed depth estimation functions and losses can often be used for both dense, and object-level depth estimation. \todo{too strong statement?}

In the case of depth prediction and corresponding loss functions, we distinguish between continuous regression approaches in contrast to approaches that rely on discretization of the depth estimation.
\subsubsection{Continuous Regression}
The reverse Huber loss (berHu) is employed in \cite{laina2016deeper} to model the value distribution of depth predictions as a continuous regression problem for dense depth estimation.
\cite{ghosh2022airtrack} uses the L2 loss for training a continuous, object-level depth regressor. The log-distance is used within the loss calculation to scale larger distance values.

\subsubsection{Depth Discretization}
\cite{cao2017estimating} formulates depth estimation as a classification problem by discretizing the depth values into intervals.
The cross-entropy (CE) loss is used to train a classifier that assigns a depth bin to every pixel for dense depth estimation.
\cite{li2018deep,li2018monocular} use a soft-weighted sum inference strategy to compute final depth predictions based on the softmax predictions scores of the depth bins.
\cite{fu2018deep} proposes an ordinal regression loss function to learn a meaningful inter-depth-class ordinal relationship.
The depth intervals for discretization are growing in width for increasing distance to the camera as the uncertainty about the true depth increases as well.
\cite{diaz2019soft} extends on the idea of using ordinal regression for depth estimation by using a softmax-like function to encode the target vector for depth classification as a probability distribution.
%

% We outline the mathematical foundation of the identified baseline approaches in more detail in \Cref{sec:methodology}.

%% file: chapters/3_methodology.tex
\section{Methodology}
\label{sec:methodology}
In this section, we give information on YOLOX as the selected base framework for 2D object detection.
We outline the mathematical foundation of the different depth estimation strategies and embed our proposed methods into these paradigms.
In addition, we introduce the Fitness score metric in detail.
%\todo{As a reader I am a bit lost in this section. What is this trying to tell me? Am I reading the methodology of a new approach? Why is this information included in the paper? it feels a bit like reading a book about this topic, rather than a conference paper}

\subsection{YOLOX -- Base Framework for 2D Object Detection}

To tackle the problem of depth estimation at the object level, we start with a pure 2D object detection framework that outputs a confidence score, class label, and 2D bounding box for each object using axis-aligned rectangles.
Given our use case, in which the trade-off between inference speed and detection performance is of high importance, we choose YOLOX Tiny \cite{yolox2021} as the base object detection framework.
YOLOX was released in 2021 and is one of the latest advances within the YOLO family \cite{redmon2016you,redmon2017yolo9000}.

To allow for object-level depth estimation, we create a separate new head dedicated to depth estimation.
The architecture of the depth head is based on the existing classification head with the necessary adjustments to the number of output channels in the last layer.
This separation between the depth head and the other heads allows for a modular combination of various outputs.

While we have selected YOLOX as the foundation for this work, the ideas presented in the following sections can be carried over to other modern 2D object detectors.

\subsection{Depth Regression}
\label{sec:methodology_depth_regression}

The most natural way to estimate depth $d$ is to frame it as a continuous regression problem.
In this case, the model is trained to predict a single and continuous value by minimizing the distance between the model predictions $\hat{y}$ and the ground truth target $y$.
In its simplest form, depth can be regressed directly, \ie $y = d$ and $\hat{y} = \hat{d}$.
This simple model, however, allows for negative distances, which are not meaningful in the context of monocular depth estimation.
Thus, we can use a differentiable transfer function, $g \left( x \right)$, which supports us in encoding and decoding the network output given a set of constraints.
%
%Generally, we can write for encoding the depth $d$ of an object into the corresponding network output $y$: \todo{this might be a bit too elaborate / too much explanation for a conference paper}
%\begin{equation}
%    y = g \left( d \right)  \quad\text{and}\quad \hat{y} = g \left( \hat{d} \right).
%\end{equation}
%For encoding the network output $y$ to the actual depth $d$, we use the inverse of $g$:
%\begin{equation}
%    d = g^{-1} \left( y \right)  \quad\text{and}\quad \hat{d} = g^{-1} \left( \hat{y} \right).
%\end{equation}
%with $\hat{d}$ and $d$ depicting the predicted and target depth, respectively.
%
% ReLu
To avoid negative predictions, we propose the encoding
\begin{equation}
    \label{eq:regression_relu_enc}
    g \left( x \right) = \frac{x - b}{a}.
\end{equation}
Its corresponding decoding can be calculated as:
\begin{equation}
    \label{eq:regression_relu_dec}
    g \left( x \right)^{-1} = \max \left(d_\text{min}, a \cdot x + b \right),
\end{equation}
with $a$ and $b$ being hyperparameters that allow for better control over the domain and range of the model outputs.
As $g^{-1}$ follows the ReLU structure, we refer to this approach as the ReLU-like encoding.
We argue that designing a differentiable transfer function with this constraint not only eases training but also enhances robustness against out-of-distribution objects \cite{blei2022identifying} or adversarial attacks.

Besides direct regression, there are encodings based on non-linear transfer functions \eg, the inverse $g \left( x \right) = \tfrac{1}{x}$ \cite{gahlert2020single} and the logarithm $g \left( x \right) = \log{x}$ \cite{eigen2014depth,eigen2015predicting,lee2019big,bhat2021adabins,ghosh2022airtrack}.
All previously mentioned encodings lack an upper bound.
Thus, they allow for any positive number to be predicted as the depth of the object.
As a result, the calculated loss is also unbound and may cause instabilities during training.
In some use cases, however, it is possible to assign an upper bound or a maximum distance to the objects of interest.
For those settings, we propose a bounded transfer function that maps the domain $\left( d_\text{min}, d_\text{max} \right)$ to the range $\left( -\infty, +\infty \right)$:
\begin{equation}
    \label{eq:regression_sigmoid_enc}
    g \left( x \right) = \text{logit} \left( \frac{x - d_\text{min}}{d_\text{max} - d_\text{min}} \right).
\end{equation}
The corresponding decoding operation, based on the sigmoid function $\sigma$, is then calculated as:
\begin{equation}
    \label{eq:regression_sigmoid_dec}
    g \left( x \right)^{-1} = \left( d_\text{max} - d_\text{min} \right) \sigma \left( x \right) - d_\text{min},
\end{equation}
where $d_\text{max}$ and $d_\text{min}$ are the maximum and the minimum depth, respectively.
As $g^{-1}$ uses the sigmoid function, we refer to this approach as the Sigmoid encoding.

\subsection{Depth Bin Classification}
\label{sec:methodology_depth_bin_classification}

Depending on the application and the use case, a coarse depth estimation might be sufficient.
In such cases, depth estimation can be framed as a multiclass classification task with $K$ discretized depth intervals $\left\{ d_0, d_1, ..., d_{K-1} \right\}$ \cite{cao2017estimating}.
Each depth interval links to an individual class in the classification paradigm.
Relaxing the need for fine-grained and continuous depth estimation also eases the process of ground truth generation and data annotation.
This, in return, can be beneficial from a business perspective.

During training, in a classification setting, the softmax function is typically used in CNNs to compute the pseudo-probability distribution and is paired with CE loss.
At test time, the selected depth bin is obtained by using the argmax over the pseudo-probabilities.
Reformulating depth estimation as a simple classification task is straightforward.
In our experiments, we will use this approach as the baseline for classification-based approaches.

Employing CE, however, models the classes -- and thus the depth bins -- as being independent of each other.
In particular, the default CE loss doesn't penalize predictions more if they are further away from the target bin compared to predictions that are closer to the target.

Depth bins, however, are ordered.
We naturally would consider predictions that are far away from the actual depth as \emph{more wrong} compared to closer ones.
% \todo{The following is an important contribution, but i think in this section it needs to be more clearly separated: what is baseline and what is new from us?}
Thus, we propose to design a loss that considers the distance of the predicted depth bin to the target depth bin.

Designing a loss based on the distance between the prediction and ground truth implies the knowledge of the argmax of the predicted depth classes.
Once the argmax and ground truth is known, an arbitrary distance loss function \eg, Smooth L1 or MSE, can easily be computed.
The implementation of this approach, however, renders a challenge as the default argmax function is not differentiable.
Thus, we replace it with the Soft-Argmax \cite{finn2016deep,goroshin2015learning}
\begin{equation}
    \text{Soft-Argmax} \left( \hat{y}, \beta \right)
    =\sum _{i=0}^{K-1} i \cdot \text{softmax} \left( \beta \hat{y} \right)_{i}
\end{equation}
where $\beta > 0$ is a parameter that scales the model predictions $\hat{y}$.
The larger the $\beta$, the more it approximates a one-hot encoded argmax.
In our experiments, we found $\beta = 3$ to be a good choice.
Soft-Argmax provides an approximated bin index that is used to compute a distance loss between it and the target bin.

During inference, we can naturally obtain the predicted depth bin, $\hat{d_i}$, by applying the argmax function to the model output, $\hat{y}$, and set the depth value, $\hat{d}$, to its center.

\subsection{Fitness Score}
\label{sec:methodology_metrics}

As described previously, depth estimation can be formulated as a regression or a classification task.
A natural choice for a metric capable of assessing the quality of depth estimation is the mean absolute localization error \cite{gahlert2020cityscapes}.
% It is a continuous number representing the absolute distance between ground truth and prediction.

If depth estimation is framed as a classification task, the predicted depth is by default not a continuous number and depends on the \emph{real} depth assigned to this bin \eg, its center.
As a result, predictions might cause a large absolute localization error despite being assigned to the proper depth bin.
This effect makes it difficult to compare both regression and classification models.

To solve this challenge, we suggest to also discretize the network prediction in the case of a regression model into $K$ bins and apply a metric suitable for classification tasks.
By doing so, we are able to compare both regression and classification models.
Finally, this approach also simplifies the proper model selection by a single metric across the different depth estimation paradigms.

As the network predicts confidence, class label, bounding box parameters as well as the depth, we effectively have set up a multitask network.
Thus, we need to be able to assess both depth estimation as well as standard 2D object detection performance.
Assessing the performance of a multitask network is challenging as all included tasks might perform and be weighted differently.
We, however, favor a single number summarizing the model performance in all tasks.

In typical object detection benchmarks, mean Average Precision (mAP) is commonly used \cite{everingham2010pascal}.
mAP measures both the classification performance as well as the bounding box localization quality by utilizing the Intersection-over-Union (IoU) of the predicted and the ground truth bounding box as an auxiliary metric.
In addition, F1 jointly assesses both precision and recall in a single number.
Note that because several properties of the object -- \eg its size and full 3D bounding box -- are unknown and thus not needed for our use case, metrics commonly used in 3D object detection -- \eg $\text{AP}_\text{3D}$ \cite{geiger2012we} and $\text{DS}$ \cite{gahlert2020cityscapes} -- are not suitable.

As we propose to calculate the performance of depth estimation as a classification task, we suggest employing a scheme similar to F1 for depth estimation as well.
% By design, mAP is a classification metric, \ie, it assesses the ability to correctly classify objects.
% For object detection, however, mAP measures both the classification performance as well as the bounding box localization quality by utilizing Intersection-over-Union (IoU) of the predicted and the ground truth bounding box as an auxiliary metric.

% The scales of these two groups of metrics are typically very different.
% For example, the domain of mAP is $\left[ 0, 1 \right]$, while the domain of RMSE is $\left[ 0, +\infty \right)$.
% Combining metrics with vastly different domains is not a trivial task and it often leads to one metric being favored.

% Instead, we decide to use classification metrics for evaluating both tasks.
% For the object detection task, there's no deviation from the state-of-the-art.
% However, to evaluate depth estimation with classification metrics we need to first transform the predicted and target depth values into corresponding labels such that we can compute true positives (TP), false positives (FP), false negatives (FN), and other metrics based on the former.
% We propose to resort to discretization, described earlier in section \ref{sec:methodology_depth_bin_classification}, and split the depth interval $\left(d_\text{min}, d_\text{max}\right)$ into a set of $K$ discrete bins $\left\{ d_0, d_1, ..., d_{K-1} \right\}$.
% By combining the two individual metrics for object detection and depth estimation, we are now able to define a single metric that summarizes the model performance.
Eventually, we calculate the joint measure as the harmonic mean between the mean $\text{F}_1$-Score of the object detector, $\text{mF}_1^\text{OD}$, and the mean $\text{F}_1$-Score of the depth estimation, $\text{mF}_1^\text{DE}$, given the detected objects.
As the harmonic mean between two numbers is the same as the $\text{F}_1$-Score, we refer to this metric as $\text{F}_1^\text{Comb}$.

The mean $\text{F}_1$-Scores for both object detection as well as for depth estimation are dependent on the confidence threshold $t_c$ as well as on the minimum required IoU threshold $t_\text{IoU}$.
It is $t_c \in \lbrace 0.00, 0.01, ..., 0.99, 1.00 \rbrace$.
All predictions with confidence below this threshold will be discarded.
For $t_\text{IoU}$, we obtain the values according to \cite{lin2014microsoft} such that $t_\text{IoU} \in \lbrace 0.50, 0.55, ..., 0.90, 0.95\rbrace$.
Predictions with an IoU $\ge t_\text{IoU}$ will be treated as TP. Predictions with an IoU $< t_\text{IoU}$ are assumed to be FP.

Finally, it is
\begin{align}
    \text{mF}_1^\text{OD}  & = \text{mF}_1^\text{OD}(t_c, t_\text{IoU})                                                                         \\
    \text{mF}_1^\text{DE}  & = \text{mF}_1^\text{DE}(t_c, t_\text{IoU})                                                                         \\
    \text{F}_1^\text{Comb} & = \text{F}_1^\text{Comb}(t_c, t_\text{IoU})                                                                        \\
                           & =\frac {2 \cdot \text{mF}_1^\text{OD} \cdot \text{mF}_1^\text{DE}}{\text{mF}_1^\text{OD} + \text{mF}_1^\text{DE}}.
\end{align}
The domain of the combined score $\text{F}_1^\text{Comb}$ is $\left[ 0 , 1 \right]$ with higher values representing a better combined performance.
As the combined score still depends on both $t_c$ and $t_\text{IoU}$, we distill it into a single value, which we refer to as \emph{Fitness} score.
We define it as the maximum of the combined score over all confidence and IoU thresholds,
\begin{align}
    \text{Fitness} & = \underset{t_c, t_\text{IoU}}{\max} \text{F}_1^\text{Comb}.
\end{align}
By doing so, we are able to assess the model performance \emph{as is} when productively deployed.

%% file: chapters/4_experiments.tex
\section{Experiments}
\label{sec:experiments}

To demonstrate the effectiveness of the proposed methods for long range object-level monocular depth estimation we design several experiments and compare them using the Amazon Airborne Object Tracking (AOT) dataset.

We split the experiments into three major groups: regression, bin classification, and ordinal regression.
Each group formulates the depth estimation task differently, implying different network architectures and loss functions.
Eventually, we evaluate the performance of each experiment.
We use 2D mAP as well as the mean absolute localization error (MALE) as individual metrics for object detection and depth estimation, respectively.
In addition, we assess the quality of each tested approach using the joint Fitness Score.

\subsection{Dataset}
\label{sec:experiments_dataset}

The Amazon AOT dataset was introduced in 2021 as part of the Airborne Object Tracking Challenge \cite{aicrowd2021aot}.
It contains a collection of in-flight images with other aircraft flying by as planned encounters.
Planned objects are annotated with a 2D bounding box (in pixels), the object label, and the distance (in meters) from the camera to a specific object.
As the metadata only contains the euclidean distance from the camera without splitting it into $x$, $y$, and $z$, we use the terms \emph{distance} and \emph{depth} interchangeably within this study.
Additionally, the sequences may contain encounters with unplanned objects.
Those objects are annotated with bounding box parameters and their specific class label -- the distance, however, is unknown.

While most other datasets that feature object-level depth annotations mostly focus on autonomous driving and only contain objects up to \SI{200}{\meter} \cite{geiger2012we,gahlert2020cityscapes,sun2020scalability,caesar2020nuscenes,janai2020computer,ye2022rope3d}, the AOT dataset features objects up to several hundreds of meters.
In our experiments, we use the \emph{partial} dataset as provided by the authors.
This subset contains objects up to \SI{700}{\meter} away from the camera.
We have observed that some objects, specifically with a range below \SI{10}{\meter}, are mislabeled with respect to the object's distance.
Thus, we removed the range annotation for these objects but kept the bounding box and classification labels so that they can still be used for training the object detector.
The images in the flight sequences are collected at a rate of \SI{10}{\hertz}.
As such, many of the images tend to be quite similar.
With the goal of shortening training time without significant degradation of performance, we use only every 5th image of this dataset.
This subset is equivalent to \SI{2}{\hertz} or \SI{20}{\percent} of the initial dataset.
We further split the flight sequences in the dataset into dedicated sets for training (\SI{60}{\percent}), validation (\SI{20}{\percent}), and testing (\SI{20}{\percent}).
By splitting the entire dataset on a sequence level, we ensure that no cross-correlations between training, validation, and test set occur.
Our selected split also provides similar distance as well as class distributions as depicted in Figures \ref{fig:depthdist} and \ref{fig:classdist}.
Sample images of the dataset including typical objects are shown in Figure \ref{fig:sampleimage}.

\input{figures/02_depth_histograms.tex}
\input{figures/01_class_histograms.tex}

\input{figures/03_sample_image.tex}
%Experiment results are reported using the \emph{test} dataset.

\subsection{Experimental Setup}

Our models are trained on $2464 \times 2464$ \si{\px} images.
We upsample and slightly stretch the original images with a resolution of $2448 \times 2048$ \si{\px} to the target resolution as the network requires squared input images with dimensions multiple of \num{32}.
We use Stochastic Gradient Descent (SGD) as our optimizer and combine it with a cosine learning rate scheduler with warm-up \cite{yolox2021}.
In total, we train for \num{15} epochs with a batch size of \num{14} using \num{2} Nvidia RTX3090 GPUs.

As described previously, our network architecture features a de-facto multitask setting.
Thus, we calculate our overall multitask loss function $\Loss$ as:
\begin{equation}
    \label{eq:overall_loss}
    \Loss = \Loss_{\text{OD}} + w_{\text{DE}}\Loss_{\text{DE}}.
\end{equation}
Accordingly, the detector loss function, $\Loss_{\text{OD}}$, is defined as:
\begin{equation}
    \Loss_{\text{OD}} = w_{\text{obj}}\Loss_{\text{obj}} + w_{\text{loc}}\Loss_{\text{loc}} + w_{\text{class}}\Loss_{\text{class}},
\end{equation}
with $\Loss_{\text{obj}}$ being the objectness, $\Loss_{\text{loc}}$ the localization, and $\Loss_{\text{class}}$ the classification loss.
$w_{\text{obj}}$, $w_{\text{loc}}$, and $w_{\text{class}}$ refer to the corresponding balancing weights.
We leave the detector loss function from YOLOX \cite{yolox2021} unchanged.
Thus, it is $w_{\text{obj}} = 1$, $w_{\text{loc}} = 5$, and $w_{\text{class}} = 1$.
We conduct experiments with different depth loss functions, $\Loss_{\text{DE}}$.
At the same time, the depth weight $w_{\text{DE}}$ is a hyperparameter.

\subsubsection{Regression}

Our first set of experiments frames depth estimation as a regression task.
As such, we set the number of output channels for the last convolutional layer of our depth estimation head to 1.
As described in section \ref{sec:methodology_depth_regression}, there are different methods of encoding depth information.
Moreover, each encoding can be combined with different distance-based loss functions.

As mentioned in section \ref{sec:experiments_dataset}, the distance of the objects to the camera is at most \SI{700}{\meter}.
Therefore, we parameterize the Sigmoid encoding such that it is defined in the domain $\left( d_\text{min}, d_\text{max} \right) \to \left( 0, 700 \right)$.

Similarly, for the ReLU-like encoding, we obtain the best results when defining the hyperparameters $a$ and $b$ in a way that it approximates the Sigmoid encoding: $a=100$ and $b=\tfrac{700}{2}$.

For the depth loss function, $\Loss_{\text{DE}}$, we use Smooth L1 (SL1) \cite{girshick2015fast} and mean squared error (MSE) loss for each encoding:

\begin{align}
    \text{SL1}(y, \hat{y}) & =
    \begin{cases}
        \frac{1}{2N} \sum_{i=1}^{N} \left( y_{i} - \hat{y_{i}} \right)^2,    & \text{if } \left| y_{i} - \hat{y_{i}} \right| \le 1 \\
        \frac{1}{N} \sum_{i=1}^{N} \left| y_{i} - \hat{y_{i}} \right| - 0.5, & \text{otherwise}
    \end{cases} \\
    \text{MSE}(y, \hat{y}) & = \frac{1}{N} \sum_{i=1}^{N} \left( y_{i} - \hat{y_{i}} \right)^2.
\end{align}
In addition, we follow \cite{laina2016deeper} and combine direct depth regression with the reverse Huber (berHu) loss:
\begin{equation}
    \text{berHu}(y, \hat{y},c) =
    \begin{cases}
        \frac{1}{N} \sum_{i=1}^{N} \left| y_{i} - \hat{y_{i}} \right|,                    & \text{if } \left| y_{i} - \hat{y_{i}} \right| \le c \\
        \frac{1}{N} \sum_{i=1}^{N} \frac{\left( y_{i} - \hat{y_{i}} \right)^2 + c^2}{2c}, & \text{otherwise.}
    \end{cases}
\end{equation}
$\hat{y}$ refers the model prediction and $y$ is the target.
$c$ is a pseudo-constant that is originally calculated as a function, $c(y, \hat y)=\tfrac{1}{5}\underset{i}{\max} \left( \left| y_{i} - \hat{y_{i}} \right| \right)$ \cite{laina2016deeper}.
$N$ refers to the overall number of predictions.

\subsubsection{Bin Classification}
\label{sec:exp_bin_classification}

The second set of experiments models depth estimation as a classification task.
The depth interval $\left( d_\text{min}, d_\text{max} \right) \to \left( 0, 700 \right)$ is uniformly discretized into $K=7$ bins with a uniform bin width of \SI{100}{\meter}.
Choosing the proper bin size is rather subjective and highly dependent on the use case.
For our use case, we find that \SI{100}{\meter} is suitable since the environment is less cluttered and objects are found at larger distances when compared to other similar applications, \eg autonomous driving, where smaller bin sizes might be desired.
Similarly, and in agreement with \cite{miclea2021monocular}, we choose uniform discretization over a log-space discretization strategy because the latter increases bin sizes at larger distances where most objects are found.
Moreover, for our use case, early detections are beneficial as we want to avoid entering other objects' airspace.

To allow the model to predict $K$ depth bins, we change the number of output channels in the last convolutional layer of our depth estimation head to $K$.

Our baseline experiment in this group uses softmax (\cf section \ref{sec:methodology_depth_bin_classification}) as the final activation and CE as the loss function.
%In total, we design four experiments that employ both the proposed Soft-Argmax (SA) as well as a combination of SA and CE:
In total, we design two experiments that employ the proposed Soft-Argmax (SA) with Smooth L1 and MSE loss.

%\begin{enumerate}
%    \item SA/SL1: $\Loss_{\text{DE}}=\text{SL1}(y, \hat{y})$
%    \item SA/MSE: $\Loss_{\text{DE}}=\text{MSE}(y, \hat{y})$
%    \item CE \& SA/SL1: $\Loss_{\text{DE}} = w_{\text{CE}} \cdot \text{CE}(y, \hat{y}) + w_{\text{SL1}} \cdot \text{SL1}(y, \hat{y})$
%    \item CE \& SA/MSE: $\Loss_{\text{DE}} = w_{\text{CE}} \cdot \text{CE}(y, \hat{y}) + w_{\text{MSE}}\cdot  \text{MSE}(y, \hat{y})$. \todo{As the experiments combining CE and SA do not outperform SA/SL1 I think we should remove them. It's never stated why we do these experiments anyway...}
%\end{enumerate}
%$y$ and $\hat y$ correspond to the predicted and target bins, respectively.
%In the experiments that combine both CE and SA, it is $w_{\text{DE}} = 1$ from Equation \ref{eq:overall_loss} as we use $w_{\text{CE}}$, $w_{\text{SL1}}$, and $w_{\text{MSE}}$ to balance the loss functions instead.

\subsubsection{Ordinal Regression}

In our last set of experiments, we follow the guidelines of \cite{fu2018deep}, framing depth estimation as an ordinal regression problem.
First, we uniformly discretize the depth into 7 bins, as previously described.
The number of output channels in the last convolution layer is set to $2\cdot \left( K -1 \right)$, where the number of bins, $K$, equals 7.
Finally, we reimplement the proposed loss function, applying it to objects instead of pixels.

\subsubsection{Metrics}

We evaluate the performance of the experiments based on different metrics.
The Fitness score proposed in Section \ref{sec:methodology_metrics} is our primary metric.
To compute it, depth is once again uniformly discretized into 7 bins with a width of \SI{100}{\meter}.
During training, we search for hyperparameters that maximize the Fitness score on the validation dataset.
Once optimal hyperparameters are found, we evaluate on the \emph{test} set and report the Fitness score as our primary metric.

Additionally, we report secondary metrics including 2D mAP with \num{10} IoU thresholds $t_\text{IoU} \in \lbrace 0.50, 0.55, ..., 0.90, 0.95\rbrace$ and the mean absolute localization error.
We furthermore evaluate the performance \wrt the number of parameters, GFLOPs, inference, and post-processing times, allowing us to compare the methods in terms of computational constraints.

\subsection{Results}

Table \ref{tab:results} summarizes the experiment results on the \emph{test} set.
Within the depth regression methods, the proposed Sigmoid encoding outperforms all other encodings.
The ReLU-like encoding performs worse compared to the Sigmoid encoding but is still competitive with the best encoding from the state-of-the-art, the logarithm.
The combination Sigmoid/SL1 performs best within this group.

\input{tables/00_results.tex}

Within the classification methods, we observe that the proposed loss functions based on Soft-Argmax perform better than the baseline with CE loss.
We obtain the best results \wrt Fitness by combining Soft-Argmax with Smooth L1 loss.
Ordinal regression also outperforms the classification with CE loss.
Our results are consistent with the results of \cite{fu2018deep}.
Overall though, it is outperformed by our proposed loss functions based on Soft-Argmax.

Table \ref{tab:results} also shows that in most experiments, extending YOLOX with an additional depth head slightly degrades the base 2D performance by means of 2D mAP.
There are notable exceptions, the combination SA/SL1 is one of them.

While the combination of Soft-Argmax and Smooth L1 loss performs best \wrt the Fitness score and 2D mAP, it doesn't yield the lowest absolute localization error.
This can easily be understood as we select the middle point of the predicted bin as the actual distance of the object, \cf Section \ref{sec:methodology_depth_bin_classification}.
In particular, Table \ref{tab:results} shows that models using the Sigmoid and the ReLU-like encoding regression perform better in this aspect.

We attempt to further improve absolute localization in the classification setting by using bin interpolation, as a postprocessing step, instead of simply choosing the center of the bin.
Following \cite{miclea2021monocular}, we define the interpolation function $f$ as:
\begin{align}
    %f & = f \left(p \left( d_{i-1} \right), p \left( d_i \right), p \left( d_{i+1} \right) \right)                                  \\
    f \left(p \left( d_{i-1} \right), p \left( d_i \right), p \left( d_{i+1} \right) \right) & = f \left( \frac{p \left( d_i \right) - p \left( d_{i-1} \right)}{p \left( d_i \right) - p \left( d_{i+1} \right)} \right).
\end{align}
$p \left( d_{i} \right)$ refers to the probability of the predicted bin, $p \left( d_{i-1} \right)$, and $p \left( d_{i+1} \right)$ are the probabilities of the neighboring bins.
The predicted depth bin is refined using:
\begin{equation}
    \hat{d} =
    \begin{cases}
        \hat{d} - \frac{s_i}{2} \cdot \left( 1 - f\left( x \right) \right) ,           & \text{if } p \left( d_{i-1} \right) > p \left( d_{i+1} \right) \\
        \hat{d} + \frac{s_i}{2} \cdot \left( 1 - f\left( \frac{1}{x} \right) \right) , & \text{otherwise}
    \end{cases}
\end{equation}
where $s_i$ is the bin size \ie, the width, of the predicted bin $i$.

Any function $f$ must shift the predicted depth towards the previous bin if $p \left( d_{i-1} \right) > p \left( d_{i+1} \right)$, shift towards the next bin if $p \left( d_{i-1} \right) < p \left( d_{i+1} \right)$, and leave it unchanged if $p \left( d_{i-1} \right) = p \left( d_{i+1} \right)$.
We then select the following strictly monotone functions $f: \left[0, 1\right] \to \left[0, 1\right]$ depicted in Figure \labelcref{fig:interpolation}:
\begin{align}
     & \text{Equiangular \cite{shimizu2005sub}}                  & f(x) & = x,                                                                                            \\
     & \text{Parabola \cite{shimizu2005sub}}                     & f(x) & = \tfrac{2x}{x+1},                                                                              \\
     & \text{SinFit \cite{haller2010statistical}}                & f(x) & = \sin \left( \tfrac \pi 2 (x - 1) \right) + 1,                                                 \\
     & \text{MaxFit \cite{miclea2015new}}                        & f(x) & = \max \left( \tfrac 1 2 \left( x^4 + x \right), 1-\cos \left( \tfrac {\pi x} 2\right) \right), \\
     & \text{SinAtanFit\footnotemark \cite{miclea2021monocular}} & f(x) & = \sin \left( \tfrac \pi 2 \arctan \left( \tfrac {\pi x} 2 \right) \right).
\end{align}
\footnotetext{The authors of \cite{miclea2021monocular} did not name the function. We refer to it as \emph{SinAtanFit} to easily identify it.}
\input{figures/00_interpolation_figure.tex}
As shown in Table \ref{tab:interpolation}, all interpolation functions show improvements over the baseline.
SinFit and MaxFit obtain the same results and perform the best out of our selection.
Despite the improvements, it is not able to surpass the Sigmoid-encoded model.
As the interpolation is part of the postprocessing and does not change the network architecture or the predicted depth bin, both Fitness score and 2D mAP remain unchanged.
\input{tables/01_results_interpolation.tex}

\subsubsection{Runtime Comparison}

Besides the quality of the predictions, another important aspect is how the different models compare at runtime.
In Table \ref{tab:performance}, representative models are benchmarked and compared.

Compared to pure 2D object detection, the inference time increases by approx. \SI{4}{\milli\second} for all proposed methods.
This is mainly caused by the increased GFLOPs coming from the additional prediction head.

Amongst the proposed methods, GFLOPs, number of parameters, and inference speed do not vary meaningfully.
Looking at postprocessing though, we observe that the classification and ordinal regression models are slower than the regression models.
This result is expected as there are more steps involved for both in order to transform the model output into the depth value.
Moreover, classification and ordinal regression models grow in complexity with an increasing number of depth bins.
Lastly, we conclude that the cost of bin interpolation is negligible.

\input{tables/02_performance.tex}

%% file: figures/02_depth_histograms.tex
\begin{figure}[t]
    \centering
    \begin{tikzpicture}

        \definecolor{color0}{rgb}{0.12156862745098,0.466666666666667,0.705882352941177}
        \definecolor{color1}{rgb}{1,0.498039215686275,0.0549019607843137}
        \definecolor{color2}{rgb}{0.172549019607843,0.627450980392157,0.172549019607843}

        \begin{axis}[
            ybar interval,
            height=\figheight,
            %legend cell align={left},
            %legend style={fill opacity=0.8, draw opacity=1, text opacity=1, draw=white!80!black, nodes={scale=0.75, transform shape}},
            tick pos=left,
            width=0.98\textwidth,
            x grid style={white!69.0196078431373!black},
            xlabel={Distance [\si{\meter}]},
            xmin=-30, xmax=730,
            xtick style={color=black},
            y grid style={white!69.0196078431373!black},
            ylabel={Relative Occurrence [\si{\percent}]},
            ytick style={color=black},
            ymin=0,ymax=30,
            legend style={fill opacity=0.8, draw opacity=1, text opacity=1, draw=white!80!black, nodes={scale=0.75, transform shape}},
            xticklabels = {${[0,100)}$,${[100,200)}$,${[200,300)}$,${[300,400)}$,${[400,500)}$,${[500,600)}$,${[600,700)}$},
                    xticklabel style={
                            font=\scriptsize,
                            % rotate=20
                        },
                    %legend pos=outer north east
                ]
            \addplot table [x=bin, y=train,col sep=comma] {figures/02_depth_data.dat};
            \addplot table [x=bin, y=val,col sep=comma] {figures/02_depth_data.dat};
            \addplot table [x=bin, y=test,col sep=comma] {figures/02_depth_data.dat};
            \legend{Training, Validation, Test}

        \end{axis}

    \end{tikzpicture}
    \caption{Distance distribution for all objects within \emph{train}, \emph{val}, and \emph{test} training set. \label{fig:depthdist}}
\end{figure}
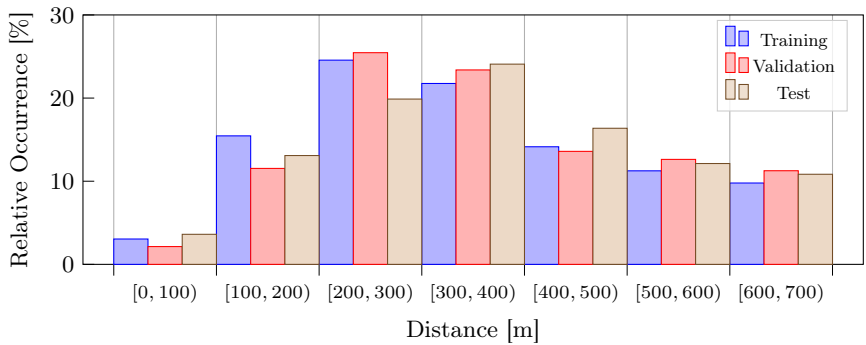

%% file: figures/01_class_histograms.tex
\begin{figure}[t]
    \centering
    \begin{tikzpicture}

        \definecolor{color0}{rgb}{0.12156862745098,0.466666666666667,0.705882352941177}
        \definecolor{color1}{rgb}{1,0.498039215686275,0.0549019607843137}
        \definecolor{color2}{rgb}{0.172549019607843,0.627450980392157,0.172549019607843}

        \begin{axis}[
                ybar interval,
                ymode=log,log origin=infty,
                height=\figheight,
                % legend cell align={left},
                legend style={fill opacity=0.8, draw opacity=1, text opacity=1, draw=white!80!black, nodes={scale=0.75, transform shape}},
                tick pos=left,
                width=0.98\textwidth,
                x grid style={white!69.0196078431373!black},
                xmin=-0.2,
                xmax=6.2,
                ymin=0.1,ymax=100,
                xtick style={color=black},
                y grid style={white!69.0196078431373!black},
                ylabel={Relative Occurrence [\si{\percent}]},
                ytick style={color=black},
                xtick = {0,1,2,3,4,5,6},
                xticklabels = {Airborne,Airplane,Bird,Drone,Flock,Helicopter},
                legend style={at={(0.615,0.8)},anchor=west}
            ]
            \addplot table [y=train,col sep=comma] {figures/01_class_data.dat};
            \addplot table [y=val,col sep=comma] {figures/01_class_data.dat};
            \addplot table [y=test,col sep=comma] {figures/01_class_data.dat};
            \legend{Training, Validation, Test}

        \end{axis}

    \end{tikzpicture}
    \caption{Class distribution for all objects within \emph{train}, \emph{val}, and \emph{test} training set, log-scaled to improve visibility. \label{fig:classdist}}
\end{figure}
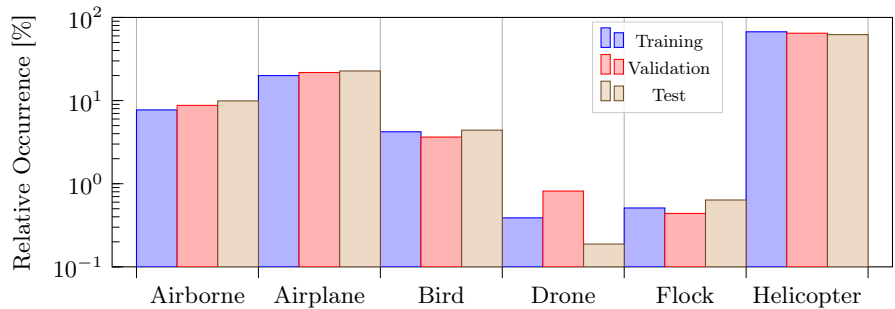

%% file: figures/03_sample_image.tex
\begin{figure}[t]
    \centering
    \includegraphics[width=0.48\textwidth]{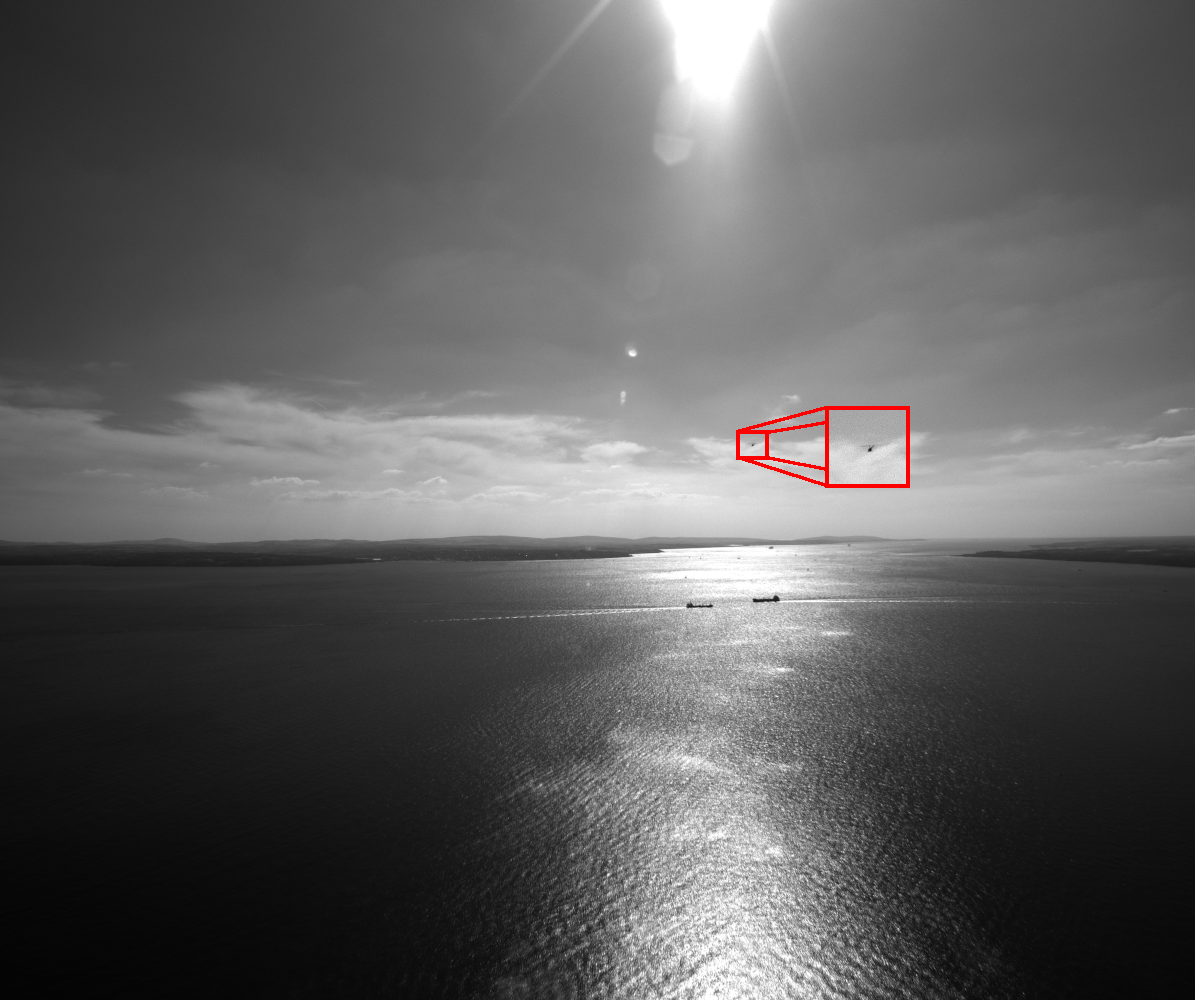}
    \includegraphics[width=0.48\textwidth]{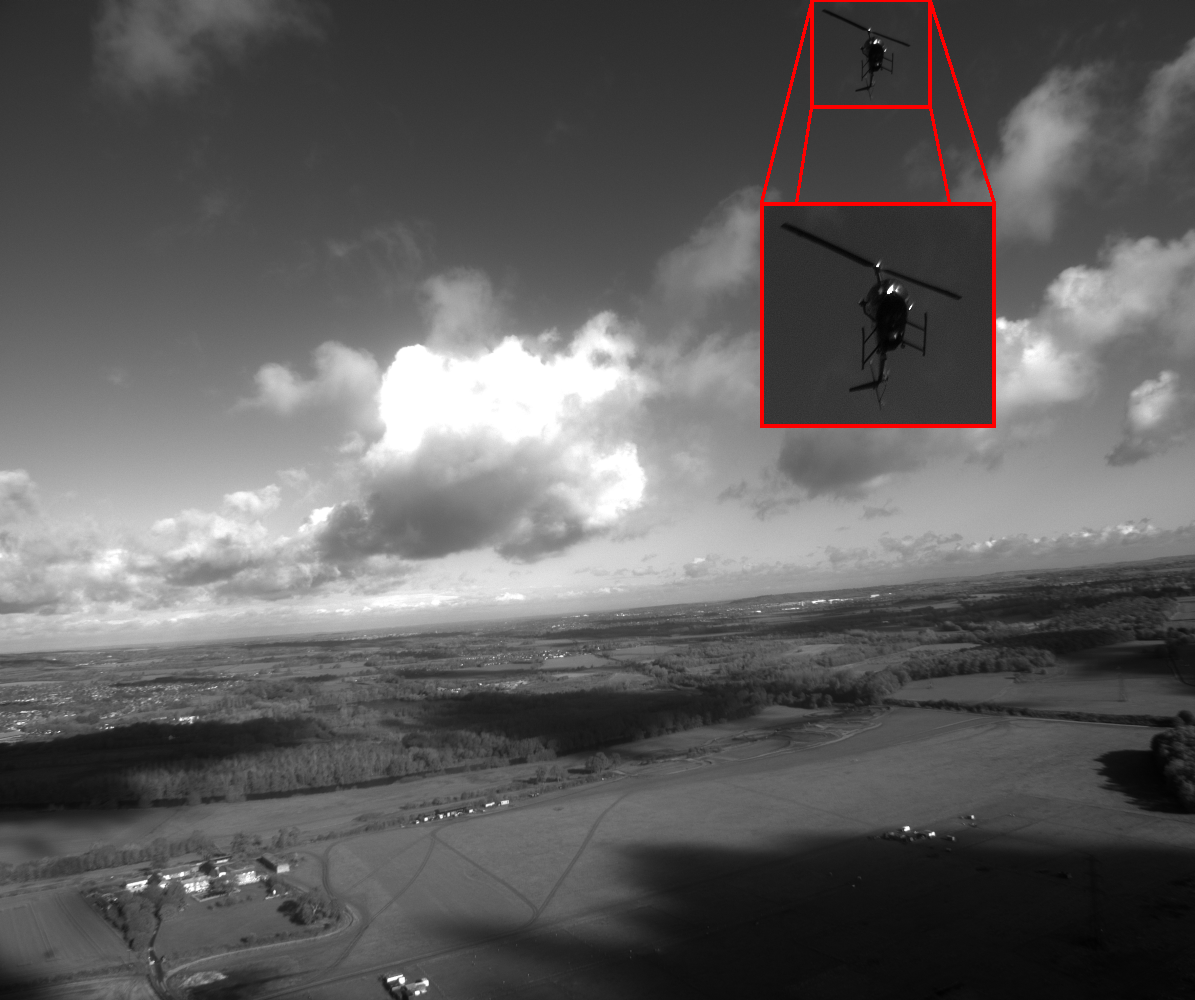}
    \caption{Sample images of the Amazon AOT dataset \cite{aicrowd2021aot}. \label{fig:sampleimage}}
\end{figure}

%% file: tables/00_results.tex
\begin{table}[t]
    \centering
    \caption{Experiment results obtained on the \emph{test} set. Object detection and depth estimation are jointly evaluated on the proposed Fitness score. 2D mAP and mean absolute localization error (MALE) individually evaluate object detection and depth estimation, respectively. \label{tab:results}}
    \begin{tabular}{l l c c c c c}
        \toprule
        \textbf{Method}                 & \textbf{Loss function}       & \textbf{Fitness}              & \textbf{2D mAP}              & \textbf{MALE}              \\ \midrule
        2D Only                         &                              & --                            & \SI{27.7}{\percent}          & --                         \\ \midrule
        \multirow{3}{*}{Direct}         & SL1                          & \SI{42.2}{\percent}           & \SI{26.7}{\percent}          & \SI{52.7}{\meter}          \\
                                        & MSE                          & \SI{43.0}{\percent}           & \SI{26.9}{\percent}          & \SI{50.3}{\meter}          \\
                                        & berHu \cite{laina2016deeper} & \SI{43.4}{\percent}           & \SI{24.7}{\percent}          & \SI{49.6}{\meter}          \\ [4px]
        \multirow{2}{*}{Inverse}        & SL1                          & \SI{39.9}{\percent}           & \SI{26.4}{\percent}          & \SI{92.8}{\meter}          \\
                                        & MSE                          & \SI{35.0}{\percent}           & \SI{25.5}{\percent}          & \SI{94.7}{\meter}          \\ [4px]
        \multirow{2}{*}{Log}            & SL1                          & \SI{48.2}{\percent}           & \SI{27.0}{\percent}          & \SI{35.5}{\meter}          \\
                                        & MSE                          & \SI{46.7}{\percent}           & \SI{26.6}{\percent}          & \SI{38.0}{\meter}          \\ [4px]
        \multirow{2}{*}{Sigmoid}        & SL1 (Ours)                   & \SI{51.6}{\percent}           & \SI{25.7}{\percent}          & \textbf{\SI{28.9}{\meter}} \\
                                        & MSE (Ours)                   & \SI{50.4}{\percent}           & \SI{25.3}{\percent}          & \SI{32.4}{\meter}          \\ [4px]
        \multirow{2}{*}{ReLU-like}      & SL1 (Ours)                   & \SI{48.3}{\percent}           & \SI{27.9}{\percent}          & \SI{33.3}{\meter}          \\
                                        & MSE (Ours)                   & \SI{47.7}{\percent}           & \SI{28.0}{\percent}          & \SI{35.5}{\meter}          \\ \midrule
        \multirow{3}{*}{Classification} & CE                           & \SI{50.9}{\percent}           & \SI{24.9}{\percent}          & \SI{37.9}{\meter}          \\
                                        & SA/SL1 (Ours)                & \textbf{ \SI{53.6}{\percent}} & \textbf{\SI{28.5}{\percent}} & \SI{37.9}{\meter}          \\
                                        & SA/MSE (Ours)                & \SI{52.8}{\percent}           & \SI{26.9}{\percent}          & \SI{38.5}{\meter}          \\ \midrule
        %& CE \& SA/SL1 (Ours)          & \SI{53.1}{\percent}           & \SI{25.8}{\percent}          & \SI{36.4}{\meter}          \\
        %& CE \& SA/MSE (Ours)          & \SI{53.5}{\percent}           & \SI{26.6}{\percent}          & \SI{37.4}{\meter}          \\
        Ordinal Regression              &                              & \SI{52.7}{\percent}           & \SI{27.0}{\percent}          & \SI{37.9}{\meter}          \\ \bottomrule
    \end{tabular}
\end{table}

%% file: figures/00_interpolation_figure.tex
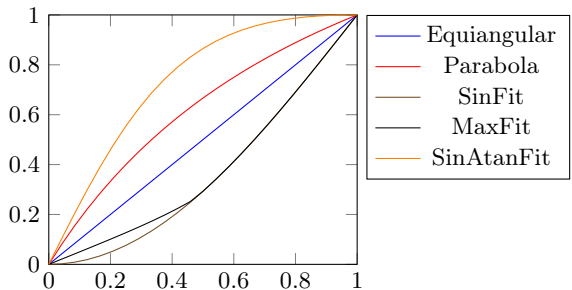
\begin{figure}[t]
    \centering
    \begin{tikzpicture}
        \begin{axis}[
                domain=0:1,
                xmin=0,xmax=1,
                ymin=0,ymax=1,
                legend pos=outer north east,
                height=\figheight]
            \addplot+[mark=none] {x};
            \addplot+[mark=none] {2*x/(x+1)};
            \addplot+[mark=none] {2*sin(pi*deg(x)/4)^2};
            \addplot+[mark=none] {max(0.5*(x^4+x),1-cos(deg(x)*pi/2))};
            \addplot+[mark=none,orange] {sin(pi/2*atan(x*pi/2))};
            \legend{Equiangular, Parabola, SinFit, MaxFit, SinAtanFit}
        \end{axis}
    \end{tikzpicture}
    \caption{Different bin interpolation functions. \label{fig:interpolation}}
\end{figure}

%% file: tables/01_results_interpolation.tex
\begin{table}[t]
    \centering
    \caption{Results of the different bin interpolation functions evaluated on mean absolute localization error (MALE). \label{tab:interpolation}}
    \begin{tabular}{l c}
        \toprule
        \textbf{Function} & \textbf{MALE}               \\ \midrule
        None (baseline)   & \SI{37.9}{\meter}           \\
        Equiangular       & \SI{31.1}{\meter}           \\
        Parabola          & \SI{32.5}{\meter}           \\
        SinFit            & \textbf{\SI{30.1}{\meter} } \\
        MaxFit            & \textbf{\SI{30.1}{\meter} } \\
        SinAtanFit        & \SI{34.6}{\meter}           \\ \bottomrule
    \end{tabular}
\end{table}

%% file: tables/02_performance.tex
\begin{table}[t]
    \centering
    \caption{Inference benchmark results on representative models for regression, classification, classification with bin interpolation, and ordinal regression. Results measured with $2464\times \SI{2464}{\px}$ image resolution, batch size 1 and FP16 using PyTorch on an Intel Core i9-10920X and Nvidia RTX3090. \label{tab:performance}}
    \begin{tabular}{l c c c c}
        \toprule
        \textbf{Method}      & \textbf{Parameters} & \textbf{GFLOPs} & \textbf{Inference}                & \textbf{Postprocessing}          \\ \midrule
        2D Only              & \num{5034321}       & \num{56.1}      & \SI{21.5}{\milli\second}          & \SI{1.5}{\milli\second}          \\ \midrule
        Sigmoid \& Smooth L1 & \num{5200690}       & \num{66.5}      & \SI{25.9}{\milli\second}          & \textbf{\SI{1.7}{\milli\second}} \\
        SA/SL1               & \num{5201369}       & \num{66.5}      & \textbf{\SI{25.4}{\milli\second}} & \SI{2.9}{\milli\second}          \\
        SA/SL1 \& SinFit     & \num{5201369}       & \num{66.5}      & \SI{25.5}{\milli\second}          & \SI{3.0}{\milli\second}          \\
        Ordinal Regression   & \num{5201951}       & \num{66.6}      & \SI{25.7}{\milli\second}          & \SI{3.0}{\milli\second}          \\ \bottomrule
    \end{tabular}
\end{table}

%% file: chapters/5_conclusion.tex
\section{Conclusion}

In this work, we addressed the problem of long range object-level monocular depth estimation and exemplarily extended the YOLOX object detection framework.
We modeled the depth estimation task as a regression, classification, and ordinal regression problem.
To jointly assess object detection and depth estimation performance, we introduced the Fitness score as a novel metric.
We proposed two novel encodings for regression, Sigmoid and ReLU-like.
The former outperforms other state-of-the-art encodings \wrt Fitness score and absolute localization error, while the latter is competitive with the best encoding from the state-of-the-art.
Moreover, for classification, we proposed a novel loss function based on the Soft-Argmax operation that minimizes the distance between the predicted and target depth bins.
In conjunction with the Smooth L1 loss, it outperforms all other models, including ordinal regression, \wrt Fitness score.
Furthermore, its 2D mAP performance even surpasses the baseline 2D model.
However, it doesn't reach the same accuracy by means of absolute localization error compared to the proposed Sigmoid encoding -- even when with bin interpolation functions.
%Compared to pure 2D detection, runtime -- inference plus postprocessing -- increases by approx. \SI{5.4}{\milli\second}, or \SI{23.5}{\percent}.
In general, regression based models have a slight advantage in postprocessing which lead to an overall faster runtime.
Based on the conducted experiments, we find that our proposed methods provide great extensions to standard 2D object detection frameworks, enabling object-level depth estimation at long range.